\documentclass{article}
\usepackage[utf8]{inputenc}

\usepackage{authblk}

\usepackage{url}
\usepackage{graphicx}
\usepackage{fullpage}
\usepackage{xcolor}
\usepackage{booktabs}
\usepackage{verbatim}
\usepackage[capitalize]{cleveref}

\usepackage{adjustbox}

\usepackage{tikz}
\usetikzlibrary{decorations.text}
\usetikzlibrary{shapes,arrows}

\usepackage{pgf}
\usetikzlibrary{arrows,automata}
\usetikzlibrary{positioning}
\usetikzlibrary{decorations.pathreplacing,angles,quotes}

\tikzstyle{line} = [draw, -latex']

\tikzstyle{root}= [rectangle,
           rounded corners,
           draw=black, very thick,
           minimum height=2em,
           inner sep=5pt,
           text centered]

\tikzset{>=latex}  

\crefname{section}{Sec.}{Secs.}
\Crefname{section}{Section}{Sections}
\Crefname{table}{Table}{Tables}
\crefname{table}{Tab.}{Tabs.}

\title{Improving Accuracy and Explainability of Online Handwriting Recognition}

\author[1]{Hilda Azimi\thanks{hilda.azimi@nrc-cnrc.gc.ca}}
\author[2]{Steven Chang\thanks{h42chang@uwaterloo.ca}}
\author[2]{Jonathan Gold\thanks{jsgold@uwaterloo.ca }}
\author[1,2]{Koray Karabina\thanks{koray.karabina@nrc-cnrc.gc.ca}}
\affil[1]{National Research Council Canada}
\affil[2]{University of Waterloo}

\newcommand{\kk}[1]{\textcolor{blue}{#1}}

\newcommand{\jg}[1]{\textcolor{orange}{#1}}

\newcommand{\etal}{\textit{et al.}}
\date{}

\begin{document}

\maketitle

\begin{abstract}

Handwriting recognition technology allows recognizing a written text from a given data. The recognition task can target letters, symbols, or words, and the input data can be a digital image or recorded by various sensors. A wide range of applications from signature verification to electronic document processing can be realized by implementing efficient and accurate handwriting recognition algorithms.
Over the years, there has been an increasing interest in experimenting with different types of technology to collect handwriting data, create datasets, and develop algorithms to recognize characters and symbols. More recently, the OnHW-chars dataset has been published that contains multivariate time series data of the English alphabet collected using a ballpoint pen fitted with sensors.
The authors of OnHW-chars also provided some baseline results through their machine learning (ML) and deep learning (DL) classifiers.

In this paper, we develop handwriting recognition models on the OnHW-chars dataset and improve the accuracy of previous models. More specifically, our ML models provide $11.3\%$-$23.56\%$ improvements over the previous ML models, and our optimized DL models with ensemble learning provide $3.08\%$-$7.01\%$ improvements over the previous DL models. In addition to our accuracy improvements over the spectrum, we aim to provide some level of explainability for our models to provide more logic behind chosen methods and why the models make sense for the data type in the dataset. 
Our results are verifiable and reproducible via the provided public repository.

\end{abstract}



\section{Introduction}
\label{s: Intro}

\subsection{Problem statement and related work}
\label{s: Related work}

Handwriting is defined to be the writing characteristics of an individual \cite{7754291}. Handwriting recognition is the process of converting written text (whether that be letters, numbers, words, symbols, etc.)  into a form which can be interpreted and recognized by a computer.  One major goal in handwriting recognition analysis involves a computer interpreting letters or words that have been written by hand. Handwriting data can be interpreted by a machine (computer) in two distinct categories: online and offline. 

Offline handwriting recognition tries to recognize a handwritten text from its static digital image that has been produced at the time of writing. Online handwriting data is captured at run time, by some sort of sensor. This sensor can be in the pen used, such as in \cite{OnHW-2020}  or through the device it was captured on a tablet or digitizer, as in \cite{LaViola}. The sensor records spatio-temporal signals in the form of a multivariate time series  throughout the process. This provides a large advantage for online handwriting over offline for certain applications or implementations.  Recording characters through a sensor, as opposed to digital images, as is in online handwriting recognition, has the added benefit of capturing more information. The sensor captures additional pieces of information such as the time it takes to write, cadence, and unique stylizations in the way it is written, all of which may not be able to be captured by a static image. These dynamic readings can produce more favourable results in certain applications as there is more information to be obtained. Applications such as signature verification \cite{FAHMY201059} and e-security or e-health \cite{faundez-zanuy_fierrez_ferrer_diaz_tolosana_plamondon_2020} can make use of  online handwriting recognition particularly well.

Over the years, there has been an increasing interest to experiment with different types of technology to collect handwriting data, to create datasets, and to develop algorithms to perform recognition of characters and symbols \cite{alpaydin_1997,wang_chuang_2012,guyon_schomaker_plamondon_liberman_janet_1994}. More recently, the OnHW-chars dataset \cite{OnHW-2020} has been published that contains multivariate time series data of the English alphabet collected using a ballpoint pen fitted with sensors. The pen used to collect this data was the STABILO DigiPen, which is fitted with various sensors: $2$ accelerometers, $1$ gyroscopes, $1$ magnetometers and a force sensor. Each letter in OnHW-chars is an $n\times 14$ vector, where $n$ is the amount of timesteps collected. For each timestep, $14$ features were collected as follows: for each of the two accelerometers, gyroscopes and magnetometers, $x$, $y$ and $z$ values were collected, as well as one feature representing time and the final feature is a binary value of whether or not force was applied with the pen on the paper. The dataset OnHW-chars provides a basis to develop algorithms to correctly identify which letter is being written based on sensor data. While publishing a dataset, the authors of \cite{OnHW-2020} also provided some baseline results through their machine learning (ML) and deep learning (DL) classifiers. To our knowledge, source codes of algorithms for processing data and trained models in \cite{OnHW-2020} are not publicly available. DL classifiers, trained on the same dataset OnHW-chars, have later been improved in a more recent paper \cite{OnHW-2022}. In addition, \cite{OnHW-2022} reports on the accuracy of classifiers trained on other new datasets, such as OnHW-equations, OnHW-symbols and OnHW-words. To our knowledge, source codes of algorithms, models, and the new datasets in \cite{OnHW-2022} have not been released to public yet.

\paragraph{A brief description of OnHW-chars}
Being the most recent and publicly available online handwriting dataset with some state-of-the-art classifiers trained on it, we focus on developing classifiers for OnHW-chars.
OnHW-chars contains multiple versions of each letter from a to z, collected from $119$ unique users. Left handed writers were excluded from the dataset.

There is a separate dataset for lowercase and uppercase letters as well as a combined dataset of both lowercase and uppercase letters. Additionally, the data was pre-split in \cite{OnHW-2020} into $5$ folds of train and test split for each of the lowercase, uppercase and combined datasets. For each of these, a writer independent (WI) and a writer dependent (WD) version of the split is provided in \cite{OnHW-2020}. In WI, writer sets of the train and the test split of a fold are disjoint. If a writer appears in the training (test) data, they are not in the test (train). However, in WD, a writer's data will be split between the training and test data. All in all, this gives $6$ datasets ($3$ cases and $2$ writer dependency) further split into $5$ folds each. In all this yields $30$ datasets. When reporting on the accuracy of OnHW-chars classifiers in \cite{OnHW-2020,OnHW-2022}, they average over $5$ folds. Thus, they present classifier accuracies on $6$ datasets, which we denote by lowercase-WD (L-WD), lowercase-WI (L-WI), uppercase-WD (U-WD), uppercase-WI (U-WI), combined-WD (C-WD), combined-WI (C-WI).




\subsection{Contributions}

\begin{figure}[t]
\begin{adjustbox}{max width=1\textwidth}
\begin{tikzpicture}[node distance = 2cm, auto, align=center]
	\tikzset{every node}=[font=\large]
    \node[root,anchor=center]
    (OnHW) 
    {OnHW Dataset \\ \cite{OnHW-2020}};    

	\node[root, right of=OnHW, xshift = 4cm, anchor=center]
	(PREPROCESSED) 
	{Variable length \\ time series data};
	
	\node[root, below of=PREPROCESSED, yshift = -2cm, anchor=center]
	(TSFRESH) 
	{Fixed length \\ feature vectors};

	\node[root, right of=PREPROCESSED, xshift = 6cm, yshift=3cm, anchor=center]
	(DL) 
	{DL Models};

	\node[root, right of=PREPROCESSED, xshift = 6cm, yshift=1cm, anchor=center]
	(DLOPT) 
	{DL Models \\ Optimized};

	\node[root, right of=TSFRESH, xshift = 6cm, yshift = 2cm, anchor=center]
	(ML) 
	{ML Models \\ -- \\ Imp: $11.3\%$-$23.56\%$};

	\node[root, right of=DLOPT, xshift = 4cm, yshift=1cm, anchor=center]
	(ENSEMBLE)
	{Ensemble Learning \\ Models \\ -- \\ Imp: $3.08\%$-$7.01\%$};

	\def\myshift#1{\raisebox{1ex}}
	\path[->] (OnHW) edge[bend left=0, postaction={decorate, decoration={text along path, text align=center,text={|\rmfamily\myshift|preprocessing}}}] (PREPROCESSED);

	\path[->] (PREPROCESSED) edge[bend left=0, postaction={decorate, decoration={text along path, text align=center,text={|\rmfamily\myshift|DL, baseline}}}] (DL);
	
	\path[->] (PREPROCESSED) edge[bend left=0, postaction={decorate, decoration={text along path, text align=center, text={|\rmfamily\myshift|DL, optimized}}}] (DLOPT);

	\path[->] (PREPROCESSED) edge[bend left=0, postaction={decorate, decoration={text along path, text align=center,text={|\rmfamily\myshift|feature}}},
	postaction={decorate, decoration={text along path, text align=center,raise={-3ex}, text={|\rmfamily\myshift|extraction}}},
	postaction={decorate, decoration={text along path, text align=center,raise={-5.5ex}, text={|\rmfamily\myshift|(tsfresh~{\cite{CHRIST201872}})}}}] (TSFRESH);

	\node[right of = PREPROCESSED,xshift=0cm,yshift=-2cm](dummy){};

	\path[->] (dummy) edge[bend left=0, postaction={decorate, decoration={text along path, text align=center,text={|\rmfamily\myshift|ML, baseline}}}] (ML);

    \draw[decorate,decoration={brace,amplitude=10pt,raise=4pt},yshift=0pt]
    (PREPROCESSED.east) -- (TSFRESH.east) node [black,midway,xshift=0.8cm] {};

	\node[right of = DL,xshift=-0.3cm,yshift=-1cm](dummy1){};
	
	\path[->] (dummy1) edge[bend left=0, postaction={decorate, decoration={text along path, text align=center,text={|\rmfamily\myshift|ensemble}}}] (ENSEMBLE);
	
	\draw [decorate,decoration={brace,amplitude=10pt,raise=4pt},yshift=0pt]
    (DL.east) -- (DLOPT.east) node [black,midway,xshift=0.8cm] {};
	
\end{tikzpicture}
\end{adjustbox}
\caption{Our workflow for training machine learning (ML) and deep learning (DL) models, and utilizing the ensemble method. Feature extraction is performed using tsfresh \cite{CHRIST201872}.
Average accuracy improvements (Imp) are reported with respect to the previous ML and DL models in \cite{OnHW-2020,OnHW-2022}.
}
\label{f: Workflow}
\end{figure}
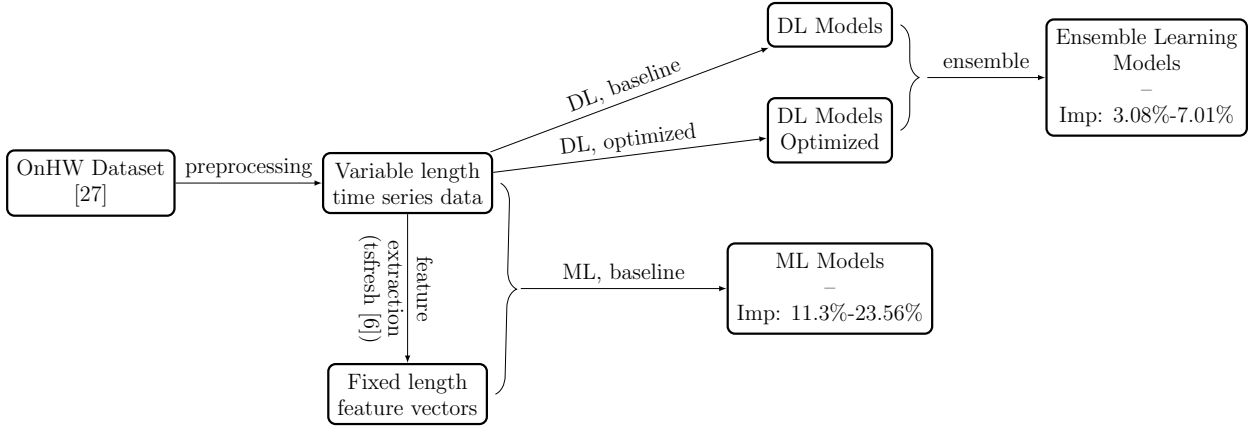

Our contributions in this paper are twofold:
\paragraph{Improving on the accuracy of classifiers on OnHW-chars}
To our knowledge, the best accuracy results of  ML classifiers and DL classifiers for the OnHW-chars dataset are the ones reported in \cite{OnHW-2020} and \cite{OnHW-2022}, respectively. As mentioned before, \cite{OnHW-2022} improves  DL classifiers in \cite{OnHW-2020}.
Our ML and DL models yield, respectively, $11.3\%$-$23.56\%$ and $2.17\%$-$4.34\%$ improvements over the accuracy of the best ML and DL models reported in \cite{OnHW-2020,OnHW-2022}.
We obtain these improvements thanks to the use of state-of-the-art feature extraction algorithms and optimizations of the models.
We further utilize the ensemble learning method to achieve $3.08\%$-$7.01\%$ improvements over the best results reported in \cite{OnHW-2020,OnHW-2022}.
We refer the reader to Figure~\ref{f: Workflow} for an overview of our workflow,
Table~\ref{tab:results} for a summary of our results, and to
Sections~\ref{s: Algorithmic aspects} and~\ref{s: Optimizations} for more details.

\paragraph{Providing verifiable and reproducible results with some level of explainability}
In addition to providing improvements to accuracy across the spectrum, this paper aims to provide some level of explainability of these models so as to provide more logic as to why these methods are chosen and why the models make sense for the type of data in the dataset. In addition to prediction accuracy, it is important to assess ML models on how they come to their decisions \cite{Fauvel2021XCM:}. One goal is to explain the reason as to why it makes sense to use the models used for the specific data at hand. For the DL models we extend the local interpretable model-agnostic explanations (LIME) \cite{ribeiro_singh_guestrin_2016} architecture to add explainability to our multivariable time series (MTS) data. We refer the reader to Sections~\ref{s: Optimizations} and \ref{s: More Explanations} for more details.

Finally, we would like to add some transparency to the process. 
Along with the paper, provided is a public repository \cite{impacx_onhw_2022} containing all the preprocessing and code for each model, so that it is reproducible and verifiable. Some models, specifically some DL models contain a level of randomness, which will alter the results slightly from one run of the model to the next. For these situations, a loadable encoded representation of the model has been provided as well. With this, we would like to make the model generation process public, so that others can have access to these models, as well as verify, reproduce, and improve our results.

\section{Processing the OnHW-chars dataset}
\label{s: OnHW dataset}

\subsection{Preprocessing and filtering}
\label{s: filter and preprocess}

The OnHW-chars dataset is provided publicly in its raw form. In \cite{OnHW-2020}, detailed explanations are provided for preprocessing OnHW-chars. However, to our knowledge, the pre-processed versions are not publicly available, which introduces some challenge for reproducing results and providing fair comparisons. According to \cite{OnHW-2020}, a high pass filter is applied to the data with a cutoff frequency of $1$~(Hz) to remove the gravitational acceleration from the accelerometer recordings. Also, a moving filter with a window of size $11$ is used and that acts as a low pass filter, allowing high-frequency noise removal from the data. It is noted in \cite{OnHW-2020} that the filtering is used when applying ML models and only trimming is applied for the DL models. Our pre-processing steps on raw data also vary depending on the choice of the classification algorithm. We will provide the details of changes in pre-processing steps for each methodology.

\begin{table}[h]
\caption{Statistical characteristics of the OnHW-chars dataset subsets in terms of mean ($\mu$) and standard deviation ($\sigma$) of the total number of timestamps for each sample of data.}
\label{tab:DataInfo}
\begin{center}
\begin{tabular}{llll}
\multicolumn{1}{c}{\bf Subset}  &\multicolumn{1}{c}{\bf $\mu$} &\multicolumn{1}{c}{\bf $\sigma$}\\
\hline
Lowercase (L) & 44.05 & 29.93\\
Uppercase (U) & 52.85 & 42.82\\
Combined (C) & 48.45 & 37.20\\
\end{tabular}
\end{center}
\end{table}

As mentioned before in Section~\ref{s: Intro}, OnHW-chars contains 6 datasets: L-WD, L-WI, U-WD, U-WI, U-WD, C-WD, and C-WI with different characteristics. Table~\ref{tab:DataInfo} presents statistical characteristics of the OnHW-chars subsets in terms of mean ($\mu$) and standard deviation ($\sigma$) of the total number of samples for each instance of data. The $\sigma$ for each subset in relation to their respected $\mu$ indicates that the number of samples for each instance in OnHW-chars is notably spread out and confirms the presence of outliers. In order to reduce the impact of outliers on the data and make the data more coherent, we discard instances based on the statistical characteristics of the lowercase subset. This is because only between $2–4\%$ of normal English literature characters are written in capitals depending on the text and the genre.
Therefore, considering real-life applications in practice, the lowercase subset can be a better candidate for detecting and removing outliers. Using $\mu$ and $\sigma$ of the lowercase subset as given in Table~\ref{tab:DataInfo}, we remove any data sample with sequence length more than $\mu+2\sigma \approx 104$. Considering that $\mu-2\sigma$ yields a negative value, we determine, based on inspecting the dataset, that any sequence of length less than $10$ should be discarded. 

\subsection{Extracting statistical features from OnHW-chars} 
\label{s: Feature extraction}

The OnHW-chars contains variable length time-series data points. On the other hand, ML algorithms commonly accept fixed length data points for their training. Therefore, for each variable-length data point in OnHW-chars, we are motivated to extract a fixed number of features that carry as much information as possible about the data.

We first pre-process OnHW-chars as mentioned in Section~\ref{s: filter and preprocess}, which is then provided as input to the tsfresh software package \cite{CHRIST201872}.
Tsfresh is the main component used in our feature extraction process. Tsfresh applies the feature extraction and scalable hypothesis testing (FRESH) algorithm \cite{christ2017distributed} to extract features from variable length time series data, and further selects (and filters) relevant features based on their significance to the classification task at hand \cite{CHRIST201872}. It computes a total of $794$ time series features, obtained from the combination of 63 different time series characterization methods. The features range from simple statistical features such as the mean, and standard deviation to more complex ones including Fourier coefficients. 

\begin{figure}[!h]
\caption{Feature extraction flowchart.}
\label{tsfresh:flowchart}
\centering
\includegraphics[width=1\textwidth]{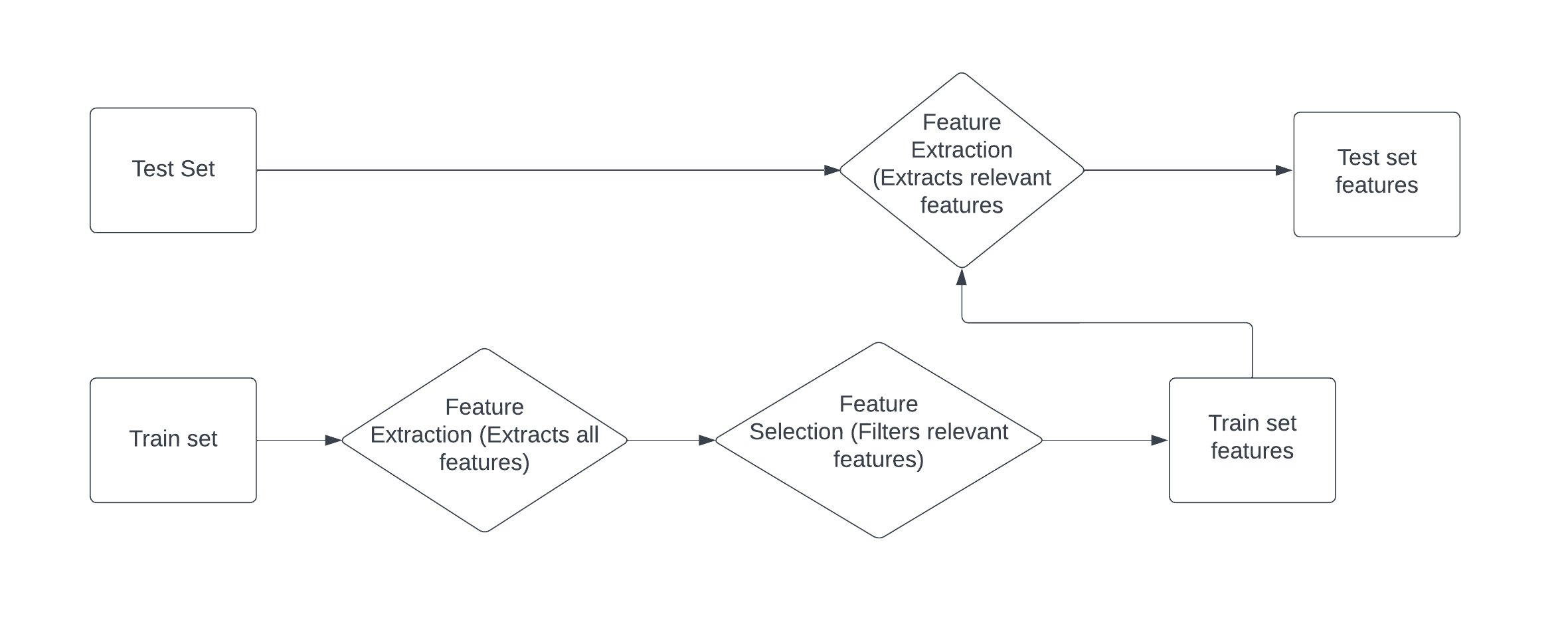}
\end{figure}

After applying tsfresh's feature extraction algorithm to OnHW-chars, there were a total of $10231$ features extracted for every letter recording. This comes as a result of $787$ features extraction per dimension for the $13$-dimensional  OnHW-chars dataset. These features are then further filtered using tsfresh's feature selection step, which conducts a series of scalable hypothesis tests and significantly drops the number of features. 
For example, $1571$ features are extracted from the first fold of L-WI in the OnHW-chars dataset.
We should emphasize that, in order to control leakage of information from the test set, we applied the feature selection step only on the training set. Once we have the selected features from the training set, those same features are directly extracted from the test set, as depicted in Figure \ref{tsfresh:flowchart}.

In our experiments, we observe that accuracy of ML classifiers is  significantly affected by the choice of the parameter n\_significant in tsfresh. The parameter n\_significant represents the number of classes a feature should be able to predict. In the case of classifying lowercase or uppercase letters, each feature we select should be able to predict any of the 26 classes. This is where the trade off comes in. When selecting n\_significant to be high, only a very small set of features will be selected. This tends to lead to low classification accuracy due to the high amount of information loss. Counter to that, when selecting n\_significant to be low, we get many features but they may not predict the majority of class labels. This trade off is later visualized in Figure~\ref{ml:knn_tuning_nsig}. Through experimentation with the k-nearest neighbors (kNN) model, we chose to use n\_significant $=17$ in our feature extraction step. We note that this choice may not be optimal and a better choice may exist. Once feature extraction and selection steps are completed, we use the extracted features to train  our ML models. During training, other transformations may be applied to data, which we explain in the following.

\section{Developing models on the OnHW-chars dataset}
\label{s: Algorithmic aspects}

\subsection{Machine learning (ML) algorithms}
\label{s: ML algorithms}

For developing ML models, we follow two strategies. In both strategies, the OnHW-chars dataset is preprocessed and filtered as described in Section~\ref{s: filter and preprocess}. In our first strategy, we extract fixed-length features from the preprocessed and filtered OnHW-chars dataset, as explained in Section~\ref{s: Feature extraction}, and that these features are provided as input to the training of ML models, where we use Decision Tree, Random Forest, Extra Trees, Logistic Regression, k-Nearest Neighbours, and Support Vector Machines. In our second strategy, we investigate the use of elastic similarity measures for classification. In particular, we use the preprocessed and filtered OnHW-chars for training kNN using Dynamic Time Warping. In the following sections, we provide details about our ML models, and we refer the reader to Table~\ref{tab:results} for our results. Table~\ref{tab:results} also provides a comparison between our results and the results from \cite{OnHW-2020}.
In particular,
we obtain
$11.3\%$-$23.56\%$ improvements over the ML models in \cite{OnHW-2020}.

\subsubsection{Strategy 1: ML models trained on fixed-length feature vectors}

As explained in Section~\ref{s: Feature extraction}, we utilize tsfresh \cite{CHRIST201872} to extract fixed-length feature vectors, which are then used as input in training some ML models. To our knowledge this is the first instance in which tsfresh  has been applied to the OnHW-chars dataset. In order to provide a fair comparison with the accuracy results as reported in Table~4 in \cite{OnHW-2020}, we train Random Forest, Decision Tree, Logistic Regression, Linear Support Vector Machines, and k-Nearest Neighbor algorithms. To further improve accuracy, we implement metric learning as described in \cite{weinberger2009distance_metric_learning} for kNN. Additionally, we train the Extra-Trees classifier implemented using scikit-learn \cite{scikit-learn}, based off of \cite{geurts2006ExtraTrees}, and SVM with nonlinear kernels. Our implementation mainly uses the software package scikit-learn \cite{scikit-learn}. In the following, we provide an overview and explanation for each algorithm we implement following Strategy~1.

\paragraph{Decision tree:}
Decision Trees are tree-like structures where each branch represents a test on a feature, and the leaf nodes represent classes. Starting at the root node, we perform a sequence of tests to move along the tree. The prediction is based on the leaf node we end up on.
To establish a baseline comparison with the results in \cite{OnHW-2020}, we use decision trees with default parameters as in scikit-learn \cite{scikit-learn}, and no additional preproccessing is applied on the extracted feature set.

\paragraph{Random forest:}
Random Forests are a collection of individual decision trees. This collection operates together as an ensemble where each tree gets a vote for the end label/class. The class with the most votes is the one that is chosen in the end.
As before, we would like a baseline comparison with \cite{OnHW-2020} to outline the effects tsfresh had on the end classification accuracy. We choose scikit-learn's default parameters: $100$ trees, no defined max depth, minimal sample split of $2$, and minimal samples leaf of $1$.

\paragraph{Extra trees:}
Extra Trees outlined in \cite{geurts2006ExtraTrees} is implemented in scikit-learn \cite{scikit-learn}, and they are similar to a Random Forests with a few key differences. These are not implemented in \cite{OnHW-2020}. We implement them with the default parameters of scikit-learn, similiar to the above mentioned models. This gives us $100$ trees, no defined maximum depth, a minimum sample split of $2$ and no specified max leaf nodes. 

\paragraph{Logistic regression:}
Logistic Regression is also a baseline we would like to establish to further outline the effects tsfresh has on the end classification accuracy. First the QuantileTransformer \cite{scikit-learn} is fitted to the selected feature set with $1000$ quantiles and a uniform output distribution. After fitting the transformer it is applied to both the test and train sets to scale each of the features. When running logistic regression we choose default parameters as in scikit-learn \cite{scikit-learn}, except that we changed the default maximum iterations parameter from $100$ to $1000$. 

\begin{figure}[!t]
\caption{kNN accuracy with various levels of n\_significant, using lowercase writer-independent (L-WI) data.}
\label{ml:knn_tuning_nsig}
\centering
\includegraphics[width=0.80\textwidth]{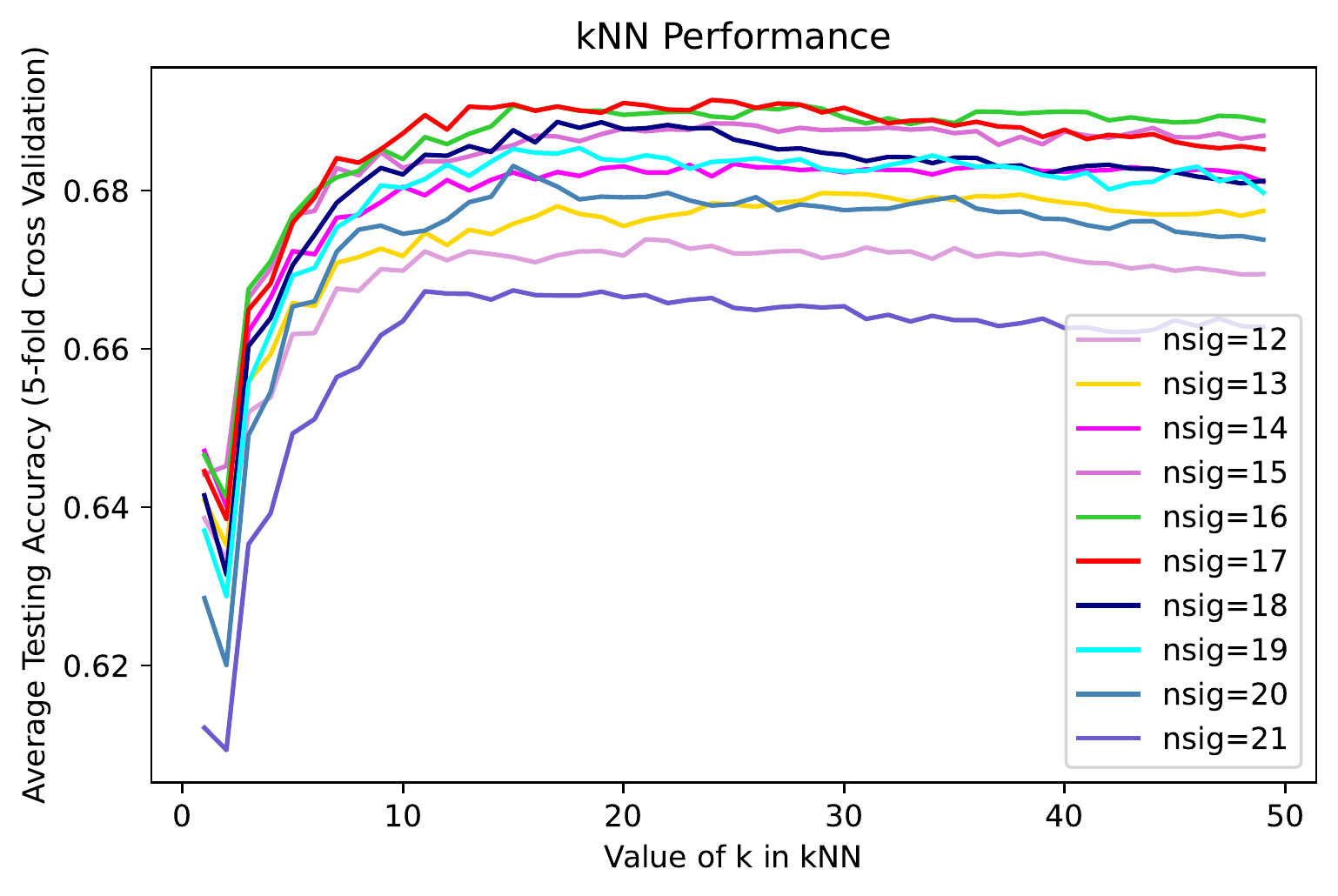}
\end{figure}

\paragraph{kNN:}
kNN is a simple non-parametric algorithm which classifies new observations based on the distance from known observations. 
We use scikit-learn's default parameter $k = 5$ to show the impact of our feature extraction method in comparison with the one in \cite{OnHW-2020}. Using the selected features from Section~\ref{s: Feature extraction}, we scale the feature set using QuantileTransformer \cite{scikit-learn} in the same way we did previously for logistic regression. No dimensionality reduction is performed. To find an appropriate level of n\_significant, we construct a series of models with varying levels of n\_significant, as seen in Figure \ref{ml:knn_tuning_nsig}. Here we take n\_significant $= 17$ for the feature selection step of all our models. 
We note that this choice may not be optimal and a better choice may exist.

\begin{figure}[t]
\caption{kNN accuracy over various levels of n\_components, using lowercase writer-independent (L-WI) data.}
\label{ml:knn_tuning_ncomp}
\centering
\includegraphics[width=0.80\textwidth]{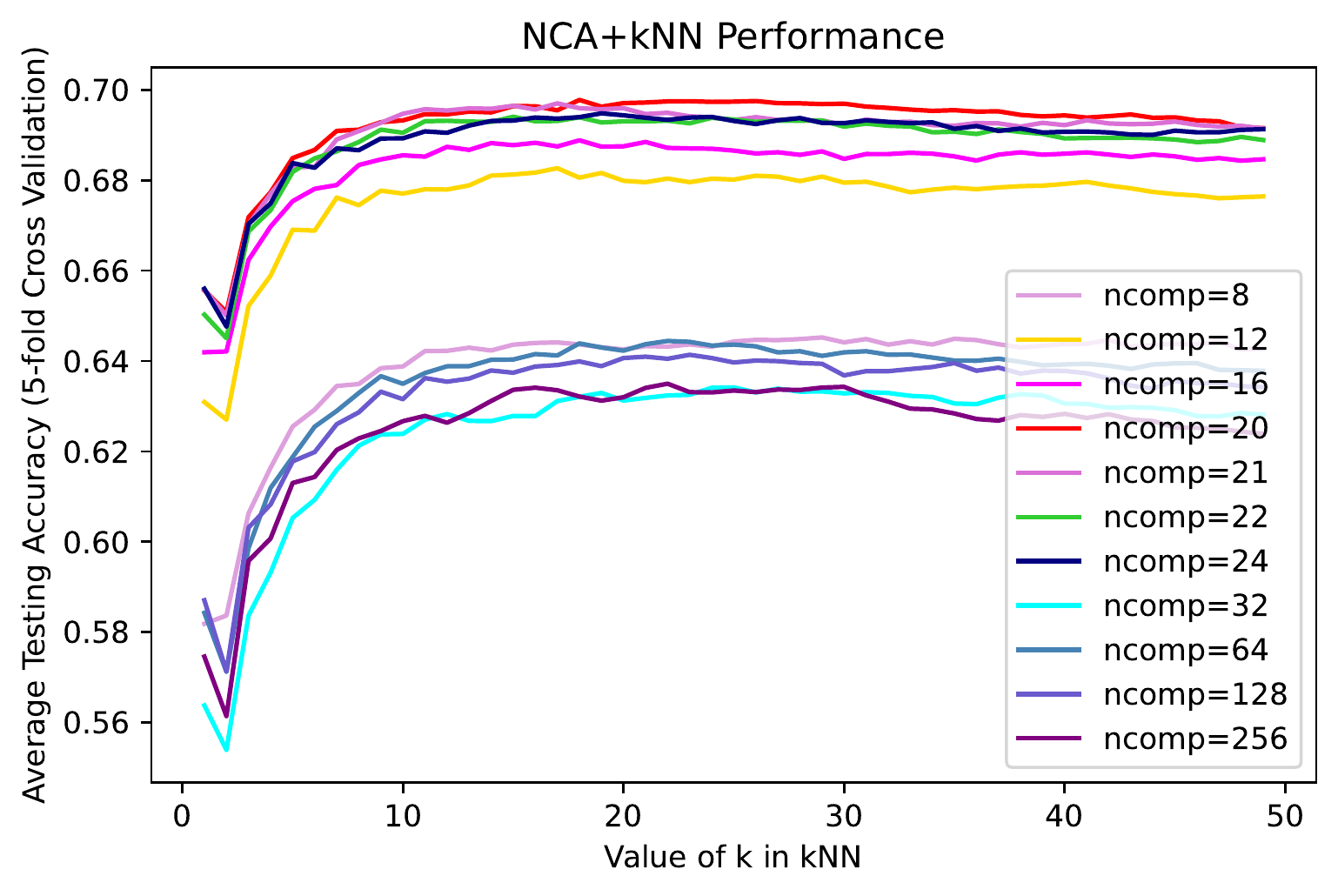}
\end{figure}

\paragraph{Supervised metric learning with kNN:}
As an extension to the previously mentioned kNN algorithm, we apply metric learning in a supervised fashion to improve accuracy scores. The pipeline remains identical, but instead of using QuantileTransformer, neighborhood component analysis (NCA) from scikit-learn \cite{scikit-learn} is implemented to learn a distance metric and reduce the dimensionality of the test and train sets. NCA improves kNN accuracy by directly maximizing the stochastic variant of the leave-one-out kNN score on the training set. When in use, NCA requires standardized data, so StandardScalar \cite{scikit-learn} was first applied to the selected features. For NCA, we specify the parameters ``init = LDA", so that LDA (i.e., linear discriminant analysis) is used to initialize the linear transformation, and ``n\_components $= 20$", reducing the inputted feature space down to $20$ features. This choice of n\_components is obtained from the accuracy of kNN models over different levels of n\_components, as seen in Figure~\ref{ml:knn_tuning_ncomp}. 

\paragraph{SVM:}
Similiar to kNN, SVM aims to group observations based on their distance from known groups. QuantileTransformer is applied to the feature set in the same fashion as we used it before. When running SVM, we used the same parameters as in \cite{OnHW-2020}. Furthermore, we implement SVM with both a linear and Gaussian kernel. 


\subsubsection{Strategy 2: ML models trained on variable-length time series data}

As opposed to some ML models that receive fixed-length feature vectors as input, some ML models can directly work with variable length time-series data. In our second strategy, we train dynamic time warping (DTW) models based on the preprocessed and filtered OnHW-chars dataset as explained in Section~\ref{s: filter and preprocess}.

\paragraph{DTW:}
DTW is a technique to measure the optimal alignment between 2 time series.  \cite{Keogh2005}. One main advantage to using DTW as a similarity measure, as opposed to Euclidean distance, is that DTW supports vectors of differing lengths. One downside to DTW, however, is that it is not a distance metric as it does not satisfy the triangle inequality and it is not positive definite \cite{Keogh2005}. Despite this, it is still often used to measure (or approximate) the distance between two time series. DTW has grown in use for time series data clustering \cite{Seto}.

Having support for differing length vectors is important for time series data and particularly crucial for sensor data such as the OnHW-chars dataset \cite{OnHW-2020}, since these vectors will often have different lengths. One approach is to zero-pad the time series and determine the distance between these zero-padded vectors using Euclidean distance, however Euclidean distance compares the points in each vector in sequence \cite{Caiado}. This is problematic for the OnHW-chars dataset because the actual length it takes to write each letter is inconsistent and slight inconsistencies in the timing will massively affect the similarity and particularly problematic when there are large differences in lengths. Variable length can also be attributed to issues with the sensor. Similarity measures which are flexible to variable length vectors, such as DTW, are preferred in this case. For our purposes, these similarity measures will be used in a kNN algorithm to classify the vectors.

\subsection{Deep learning (DL) algorithms} 
\label{s: DL algorithms}

In order to provide a comparison with the (baseline) results in \cite{OnHW-2020,OnHW-2022}, we first train DL models considered in \cite{OnHW-2020,OnHW-2022} using
the software package tsai \cite{tsai} and its default parameters. 
Our trained models include fully convolutional network (FCN),  residual neural network (ResNet), long short-term memory (LSTM), bidirectional LSTM (BiLSTM), inception time (InceptionTime), xception time (XceptionTime), and explainable convolutional neural network (CNN) for multivariate time series (XCM). In the following, we briefly introduce and explain each DL methodology that we train on OnHW-chars.
We refer the reader to Table~\ref{tab:results} for our results. Table~\ref{tab:results} also provides a comparison between our results and the best of the results from \cite{OnHW-2020,OnHW-2022}.
Our baseline models yield $2.17\%$-$3.91\%$ improvements over the DL models in \cite{OnHW-2020,OnHW-2022}.
As explained later in Section~\ref{s: Optimizations}, we further optimize 
the architectural parameters and hyperparameters of some of our DL models and utilize ensemble learning to increase the accuracy of our DL models. In the end, we obtain
$3.08\%$-$7.01\%$ improvements over the DL models in \cite{OnHW-2020,OnHW-2022}; see Table~\ref{tab:results}.
Our DL models have been developed using the software package tsai \cite{tsai}, where we used the default architectural parameters as in tsai \cite{tsai} while setting the epoch number to $50$ and the learning rate to $0.001$. We provide our source code in \cite{impacx_onhw_2022}.

\paragraph{FCN:} 
FCN has been shown to achieve state-of-the-art performance on the task of classifying time series sequences \cite{Wang2016Time}. The basic block of FCN is a convolutional layer followed by a batch normalization layer and a ReLU activation layer. The final network is built by stacking three convolution blocks. Like ResNet, the FCN architecture excludes pooling operations to prevent overfitting \cite{Wang2016Time}. In order to improve generalization, batch normalization is applied to speed up convergence speed. After the convolution blocks, the features are fed into a global average pooling layer instead of a fully connected layer, reducing the number of weights. The final label is produced by a softmax layer \cite{Wang2016Time}.

\paragraph{LSTM:}
An LSTM is a type of RNN that can learn long-term dependencies between time steps of sequential data. Contrary to CNN, an LSTM can remember the state of the network between predictions. The essential components of an LSTM network are a sequence input layer to incorporate time-series data into the network and an LSTM layer to learn long-term dependencies between time steps of sequence data. The LSTM layer contains hidden units providing inputs to memory cells and their corresponding gate units. All units (except for gate units) have connections to all units in the next layer \cite{Hochreiter1997Long}. In our implementation, we set the LSTM parameter n\_layers in tsai \cite{tsai} to $2$.

\paragraph{BiLSTM:}
The BiLSTM network extends the traditional LSTM networks. While the LSTM layer considers the time sequence in a forward direction, the BiLSTM layer considers it both backward and forwards \cite{Graves2005Framewise}. Indeed, the BiLSTM network trains two LSTM networks on the input sequence. During this process, the first recurrent layer is replicated in the network, and therefore two layers are created side-by-side. The input sequence will be an input to the first layer; meanwhile, its reversed replica will be an input to the second layer. This approach adds additional context to the network, resulting in faster and better model learning. 

\paragraph{LSTM-FCN \& BiLSTM-FCN:}
It has been shown that the performance of FCN for time series classification can enhance by adding LSTM sub-modules \cite{Karim2018LSTM}. The fully convolutional part of this architecture consists of three temporal convolutional blocks. Each block contains a temporal convolutional layer, accompanied by batch normalization followed by a ReLU activation function. Following the final convolution block, global average pooling is applied. In parallel, the time-series input is fed into a dimension shuffle layer. Next, the transformed time series as the output of the shuffle layer is passed into the LSTM block containing the LSTM layer, followed by a dropout. Finally, the output of the global pooling layer (from the fully convolutional part) and the LSTM block are concatenated and passed onto a softmax classification layer \cite{Karim2018LSTM}. 

\paragraph{MLSTM-FCN \& MBiLSTM-FCN:}
MLSTM-FCN and MBiLSTM-FCN are multivariate time series classification models whose architecture is based on the univariate time series classification models, including LSTM-FCN and Attention LSTM-FCN. The fully convolutional part in both groups (i.e., univariate and multivariate time series classification models) consists of three temporal convolutional blocks. However, compared to LSTM-FCN, the first two convolutional blocks conclude with a squeeze-and-excite block that adaptively recalibrates the input feature maps. This process can be considered as a form of learned self-attention on the output feature maps of prior layers \cite{Karim2019Multivariate}.

\paragraph{ResNet:}
Similar to FCN, the architecture of ResNet consists of three residual blocks, followed by a global average pooling layer and a softmax layer. The presence of a shortcut connection in each residual block makes the structure of Resnet very deep and enables the gradient to flow directly through the bottom layers. The filters in both architectures of FCN and ResNet are very similar. While the convolution extracts the local features in the temporal axis, the sliding filters take into account the dependencies among different time intervals and frequencies. 	Compared to FCN, ResNet is claimed to be a better candidate to be applied to larger and more complex data because it is more likely to strike a good trade-off between generalization and interpretability \cite{Wang2016Time}.

\paragraph{ResCNN:}
ResCNN is a hybrid scheme for time series classification that integrates a residual network with a CNN. In ResCNN, the strength of ResNet and CNN are combined. ResNet can learn highly complex patterns in the data due to the presence of a shortcut connection technique. However, this technique is computationally expensive and can easily cause overfitting. On the other hand, although CNN is capable of learning the temporal and spatial patterns from raw data, it cannot recover the complex patterns in the data because of few levels of the network. The architecture of ResCNN is constructed by facilitating a residual learning block at the first three convolutional layers to incorporate the strength of both networks. Additionally, batch normalization and diverse activation functions are adopted in different layers of ResCNN to enhance the nonlinear abstraction capacity. Moreover, in order to avoid overfitting, the pooling operation is removed, and the features are fed into a global average pooling instead of a fully connected layer \cite{Zou2019Integration}.

\paragraph{InceptionTime:}
Inspired by the Inception-v4 architecture, the InceptionTime is an ensemble of 5 Inception networks, with each prediction given an even weight. Each Inception network contains only two residual blocks compare to ResNet, with three residual blocks. Each block in the inception network comprises three Inception modules rather than traditional fully convolutional layers. The first principal component of the Inception module is the “bottleneck” layer which allows the Inception network to have much longer filters than ResNet (almost ten times), with roughly the same number of parameters to be learned. The second major component of the Inception module is sliding multiple filters of different lengths simultaneously on the same input time series \cite{Fawaz2020InceptionTime:}.

\paragraph{XceptionTime:}
Inspired by InceptionTime, XceptionTime is designed to be independent of the time window. The use of adaptive average pooling in this architecture makes XceptionTime more robust to the temporal translation of the inputs as the temporal information will sum out. One key difference between the XceptionTime module and InceptionTime module previously proposed in \cite{Fawaz2020InceptionTime:}, is adopting depthwise separable convolutions, which significantly mitigates the required number of parameters in the network and also can lead to higher accuracy \cite{Rahimian2019XceptionTime:}.

\paragraph{XCM:}
Compared to typical CNN architectures, XCM extracts observed variables features (2D convolution filters) and time features (1D convolution filters) directly from the input data. Features related to time fully incorporate the timing information from the input data, not from the processed features related to observed variables (features maps from 2D convolution filters). Therefore, on average, this process can lead to a better classification performance than the 2D/1D sequential approach. XCM uses 1D global average pooling followed by a softmax layer for classification, which reduces the number of trainable parameters and improves the network's generalization ability \cite{Fauvel2021XCM:}.


\begin{table}[t]
\centering
\caption{A summary of our results in comparison with \cite{OnHW-2020,OnHW-2022}. Columns labeled with ``OnHW" show the best of best results from Table~4 in \cite{OnHW-2020} and Table~6 in \cite{OnHW-2022}. 
Values in bold font indicate
the best result of their column in their group in the table. Values in bold font that are also underlined indicate the best result in their groups in the table. There are $5$ groups: ML Baseline, DL Baseline, DL Optimized, Ensemble Learning, and Best Overall.}

\label{tab:results}
\begin{center}
\resizebox{\textwidth}{!}
{\begin{tabular}{ll|cc|cc|cc|cc|cc|cc|}
\toprule
 &  & \multicolumn{4}{|c|}{Lowercase} & \multicolumn{4}{|c|}{Uppercase} & \multicolumn{4}{|c|}{Combined} \\
 &  & \multicolumn{2}{|c|}{WD} & \multicolumn{2}{|c|}{WI} & \multicolumn{2}{|c|}{WD} & \multicolumn{2}{|c|}{WI} & \multicolumn{2}{|c|}{WD} & \multicolumn{2}{|c|}{WI} \\
 &  & Proposed & OnHW & Proposed & OnHW & Proposed & OnHW & Proposed & OnHW & Proposed & OnHW & Proposed & OnHW \\
\midrule
{ML Baseline} & 5NN with NCA & 77 & - & 68.35 & - & 81.51 & - & 74.72 & - & 64.96 & - & 55.73 & - \\
 & 5NN with PCA & 57.7 & - & 44.42 & - & 59.49 & - & 48.7 & - & 36.1 & - & 23.5 & - \\
 & DT & 49.34 & 30.49 & 44.85 & 22.89 & 55.91 & 33.23 & 51.48 & 24.32 & 37.39 & 20.32 & 33.06 & 20.33 \\
 & ET & 70.49 & - & 61.81 & - & 73.83 & - & 66.15 & - & 56.8 & - & 48.2 & - \\
 & KNN-DTW & 65.41 & - & 54.62 & - & 72.08 & - & 62.72 & - & 50.69 & - & 41.14 & - \\
 & LogReg & 71.98 & 56.16 & 66.54 & 49.6 & 77.36 & 62.59 & 73.55 & \textbf{53.26} & 60.68 & 43.95 & 54.87 & 41.66 \\
 & RFC & 71.45 & 58.02 & 64.97 & 45.55 & 75.32 & 63.19 & 69.25 & 45.96 & 58.33 & 43.6 & 51.65 & 43.62 \\
 & Linear SVM & 75.64 & \textbf{62.09} & 68.1 & \textbf{51.8} & 80.75 & \textbf{70.61} & 74.67 & 54 & 66.24 & \textbf{48.77} & 58.04 & \textbf{46.56} \\
 & RBF SVM & \underline{\textbf{78.42}} & - & \underline{\textbf{71.08}} & - & \underline{\textbf{81.91}} & - & \underline{\textbf{76.82}} & - & \underline{\textbf{66.89}} & - & \underline{\textbf{59.87}} & - \\
 & KNN & 67.61 & 49.17 & 55.96 & 34.09 & 70.29 & 57.49 & 61.42 & 36.68 & 50.33 & 38.3 & 39.57 & 33.08 \\
\midrule
\multicolumn{2}{l|}{ML Improvements} & \multicolumn{2}{c|}{16.33} & \multicolumn{2}{c|}{19.28} & \multicolumn{2}{c|}{11.3} & \multicolumn{2}{c|}{23.56} & \multicolumn{2}{c|}{18.12} & \multicolumn{2}{c|}{13.31}\\
\midrule
{DL Baseline} & FCN & 86.93 & 81.62 & 73.67 & 71.48 & 88.16 & 85.37 & 78.71 & 77.24 & 77.39 & 67.41 & 62.15 & 58 \\
 & InceptionTime & \underline{\textbf{92.74}} &  84.14 & \underline{\textbf{83.44}} & 75.28 &  \underline{\textbf{94.54}} & 87.8 & \underline{\textbf{88.43}} & 81.62 & 83.17 & 70.43 & 70.49 & 61.68 \\
 & BiLSTM-FCN & 86.53 & - & 73.74 & - & 88.11 & - & 79.54 & - & 77.89 & - & 63.31 & - \\
 & LSTM-FCN & 86.5 & 81.43 & 74.36 & 71.41 & 88.28 & 85.43 & 79.91 & 77.07 & 77.92 & 67.34 & 63.26 & 57.93 \\
 & LSTM & 88.33 & 79.83 & 78.33 & 73.03 & 90.83 & 88.68 & 84.59 & 81.91 & 79.15 & 67.83 & 67.61 & 60.29 \\
 & BiLSTM & 88.69 & 82.43 & 78.5 & 75.72 & 91.3 & 89.15 & 84.37 & 81.09 & 79.42 & 69.37 & 67.5 & 63.38 \\
 & (Bi)MLSTM-FCN & 86.7 & - & 74.49 & - & 89.2 & - & 80.74 & - & 79.15 & - & 65.1 & - \\
 & MLSTM-FCN & 86.63 & 80.21 & 74.15 & 71.9 & 89.16 & 85.25 & 80.89 & 77.44 & 79.12 & 69.33 & 64.82 & 60.14 \\
 & ResCNN & 90.23 & 82.52 & 78.26 & 72 & 91.53 & 86.91 & 82.41 & 78.64 & 80.22 & 67.55 & 65.43 & 58.67 \\
 & ResNet & 92.48 & 83.01 & 81.64 & 71.93 & 94.11 & 86.41 & 86.28 & 78.03 & 82.84 & 68.56 & 68.65 & 58.74 \\
 & XCM & 81.94 & 74.39 & 72.36 & 68.12 & 84.11 & 81.67 & 76.41 & 74.32 & 70.99 & 58.18 & 61.82 & 51.99 \\
 & XceptionTime & 91.95 & 81.41 & 82.86 & 70.76 & 94.02 & 85.94 & 87.93 & 78.23 & \underline{\textbf{83.32}} & 66.7 &  \underline{\textbf{71.8}} & 56.92 \\
 & CNN-BiLSTM & - & \bfseries 89.66 & - & \bfseries 80 & - & \bfseries 92.58 & - & \bfseries 85.64 & - & \bfseries 78.98 & - & \bfseries 68.44 \\
\midrule
{DL Optimized} & InceptionTime & \underline{\textbf{92.79}} & - & \underline{\textbf{83.91}} & - & \underline{\textbf{94.75}} & - & \underline{\textbf{88.74}} & - & \underline{\textbf{82.71}} & - & \underline{\textbf{71.82}} & - \\
 & LSTM-FCN & 85.27 & - & 75.87 & - & 89.44 & - & 81.82 & - & 77.72 & - & 65.8 & - \\
 & LSTM & 89.49 & - & 80.86 & - & 91.32 & - & 85.26 & - & 79.16 & - & 70.51 & - \\
 & MLSTM-FCN & 87.35 & -  & 76.36 & - & 87.35 & - & 80.65 & - & 79.19 & - & 67.83 & - \\
\midrule
\multicolumn{2}{l|}{DL Improvements} & \multicolumn{2}{c|}{3.13} & \multicolumn{2}{c|}{3.91} & \multicolumn{2}{c|}{2.17} & \multicolumn{2}{c|}{3.1} & \multicolumn{2}{c|}{4.34} & \multicolumn{2}{c|}{3.38}\\
\midrule
{Ensemble Learning} & Plurality (top 3) & 93.47 & - & 84.95 & - & 94.98 & - & 89.27 & - & 85.08 & - & 74.13& - \\
 & Soft (top 2 + opt.) & 93.66 & - & 85.39 & - & 95.14 & - & 89.74 & - & 85.36 & - & 74.56 & - \\
 & Weighted Soft & \underline{\textbf{94.18}} & - & \underline{\textbf{86.12}} & - & \underline{\textbf{95.66}} & - & \underline{\textbf{90.34}} & - & \underline{\textbf{85.99}} & - & \underline{\textbf{74.8}} & -\\
 & 
 Soft (all) & 93.34 & - & 84.52 & - & 94.76 & - & 89.01 & - & 85.49 & - & 73.77 & - \\
\midrule
Best Overall &  & \underline{\textbf{94.18}} & \textbf{89.66} & \underline{\textbf{86.12}} & \textbf{80} & \underline{\textbf{95.66}} & \textbf{92.58} & \underline{\textbf{90.34}} & \textbf{85.64} & \underline{\textbf{85.99}} & \textbf{78.98} & \underline{\textbf{74.8}} & \textbf{68.44} \\
\midrule
\multicolumn{2}{l|}{Overall Improvements} & \multicolumn{2}{c|}{4.52} & \multicolumn{2}{c|}{6.12} & \multicolumn{2}{c|}{3.08} & \multicolumn{2}{c|}{4.7} & \multicolumn{2}{c|}{7.01} & \multicolumn{2}{c|}{6.36} \\
\bottomrule

\end{tabular}}
\end{center}
\end{table}

\section{Optimizations}
\label{s: Optimizations}

In the folowing sections, we discuss architectural parameter and hyperparameter optimization of some of our DL models and also the use of ensemble learning to further improve the accuracy of our DL models. 
Even though we observed similar accuracy after parameter optimizations,
the use of ensemble learning provided up to $2.98\%$ improvements over our DL models, whence $3.08\%$-$7.01\%$
improvements over the results reported in \cite{OnHW-2020,OnHW-2022}; see Table~\ref{tab:results}. For DL baseline methods in \cite{OnHW-2020,OnHW-2022}, as indicated in Table~\ref{tab:results}, CNN-BiLSTM had the best performance among all other DL models. However, we could not implement the same model due to the lack of detailed information about CNN architecture, such as the total number of layers, the total number of filters in each layer, and filter sizes. CNN, as a class of artificial neural networks, can take different structures leading to different performances and dramatically affecting the results for comparison purposes. Nevertheless, compared to CNN-BiLSTM, our InceptionTime implementation had a better performance overall.

\subsection{Optimizing acrhitectural parameters and hyperparameters}

Since the InceptionTime model did better overall among our base DL models, we decided to optimize it. Due to a large choice of architectural parameters in LSTM, we also decided to optimize LSTM, LSTM-FCN, and MLSTM-FCN. Of course, other models could be optimized but we leave this for future work. 
We used the optuna framework \cite{optuna_2019} with $100$ number of trials in our study. 
Across all of the optimization studies, our search space for the ``learning rate" is set between $0.00001$ and $0.01$ with logarithmic increments and the ``epoch number" is exhausted from $25$ to $100$ with increments of $25$. In our study, search space for InceptionTime parameters are as follows: ``nf" takes $4,8,16,32,40,48,56,64,128$; ``depth" takes values from $1$ to $15$ with increments of $1$ and ``fc\_dropout" takes values from $0$ to $0.9$ with increments of $0.1$
The search space for LSTM parameters are as follows: ``n\_layers" takes values from $1$ to $5$ with increments of $1$; ``rnn\_dropout" and ``fc\_dropout" take values from $0$ to $0.9$ with increments of $0.1$; ``bidirectional" takes $\mbox{True}$ or $\mbox{False}$. 
The search space for LSTM-FCN and MLSTM-FCN parameters are as follows:
``rnn\_layers" takes values from $1$ to $5$ with increments of $1$; ``rnn\_dropout" and ``fc\_dropout" takes values from $0$ to $0.9$ with increments of $0.1$; ``bidirectional" takes $\mbox{True}$ or $\mbox{False}$. We first optimized the model parameters independently on the L-WI, U-WI, and C-WI datasets. We used the same parameter sets when training our models on the respective WD datasets. Therefore, for a small number of cases, our models trained on WD underperform in comparison with the base DL models, where the most significant drop is observed for MLSTM-FCN trained on the U-WD dataset. One could ideally optimize the parameters independently on WD as well, but we do not pursue this path mainly because of the high cost of optimizing parameters and the minor performance gains due to parameter optimization over our base models (e.g.~compare the accuracy of InceptionTime in Table~\ref{tab:results} for DL Baseline vs.~DL Optimized). 
More details, including the optimized models and their parameters can be found in our repository \cite{impacx_onhw_2022} and the accuracy of the optimized results are presented in Table~\ref{tab:results}.
Our optimization efforts resulted in similar accuracy in comparison with our base DL models.

\subsection{Optimizing via the use of ensemble learning}

While the models above outperform the previously best published results and the optimizations of these models yield even more accurate results, we wanted to find a way to increase the accuracy even more. We begin by analyzing how the models make predictions and specifically, how each model differs from each other when making predictions. We look at the data where (the best) models make incorrect predictions and see how other models perform in these points. This analysis leads us to believe that combining the predictions of these models, in an ensemble method, may be able to help lead us to increased accuracy \cite{dietterich_2000}. We present the findings of our analysis, as well as results below.

\subsubsection{Failure space}

We will define the failure space of a classification model as the space of input data where the model makes an incorrect predictive classification. We examined how the failure spaces of each of the 16 models trained above (12 base models and the 4 optimized models), compared to one another. It was hypothesized that given the different architectures and even parameters (on a small scale) would cause different models to fail in different places. In other words, the intersection of the failure spaces of these models does not fully overlap. If this were to be the case, then there would be reason to believe that the models may be focusing on different aspects of the MTS and that there is potential for a combination of models to be used to collectively produce an output. Below, are two seperate analyses, exploring the failure spaces of the 16 models. 

\begin{figure}[t]
\centering{\includegraphics[width=0.80\textwidth]{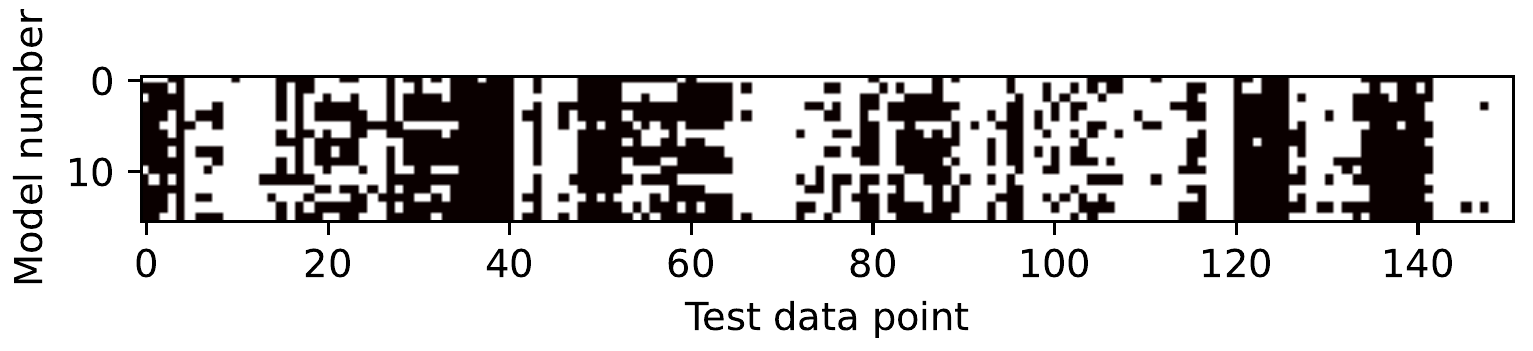}}
\caption{The prediction space analysis in the both independent fold 0 dataset. The models are numbered as follows: 0:XceptionTime,1:InceptionTime, 2: ResNet, 3: LSTM FCN, 4: BiLSTM FCN, 5:LSTM, 6:BiLSTM, 7:MLSTM FCN, 8:BiMLSTM FCN, 9:FCN, 10:ResCNN, 11:XCM, 12: Optimized InceptionTime, 13: Optimized LSTM, 14: Optimized LSTM FCN, 15:Optimized MLSTM FCN.}
\label{fig:Prediction Space}
\end{figure}

First, we explore how each of the models perform in comparison to one another. The dataset is fixed to the C-WI, fold 0 dataset. We then fix a letter, say ``Z" and examine all test data points that are actually ``Z"s. For each piece of test data, we determine whether each model predicted it correctly or incorrectly. The results are displayed in a heatmap in Figure \ref{fig:Prediction Space}. Each row corresponds to the prediction of a different model, and each column corresponds to a unique test data. The colour represents a correct or incorrect prediction, with a correct prediction being white and an incorrect prediction being black. We call this type of an analysis the prediction space analysis.

The letter ``Z" was chosen for this analysis as this was poorly predicted by the models. We find that letters which look very similar in upper and lower case tend to do poorly as they get mistaken for each other frequently. ``Z" is one example of this and was consistently one of the worst performing letters. First, we notice that the map looks very spotted. There are a few columns that are all black, but there are also many columns that are split between correct and incorrect predictions. This tells us that there may be a use to combining these models in such a way to diminish the incorrect predictions. There will be more analysis below into how models correctly and incorrectly predict data points. The prediction space analysis shows us that oftentimes, when even the best performing models are incorrect, other models may predict the data point correctly.

\begin{figure}[t]
\centering\includegraphics[width=0.6\textwidth]{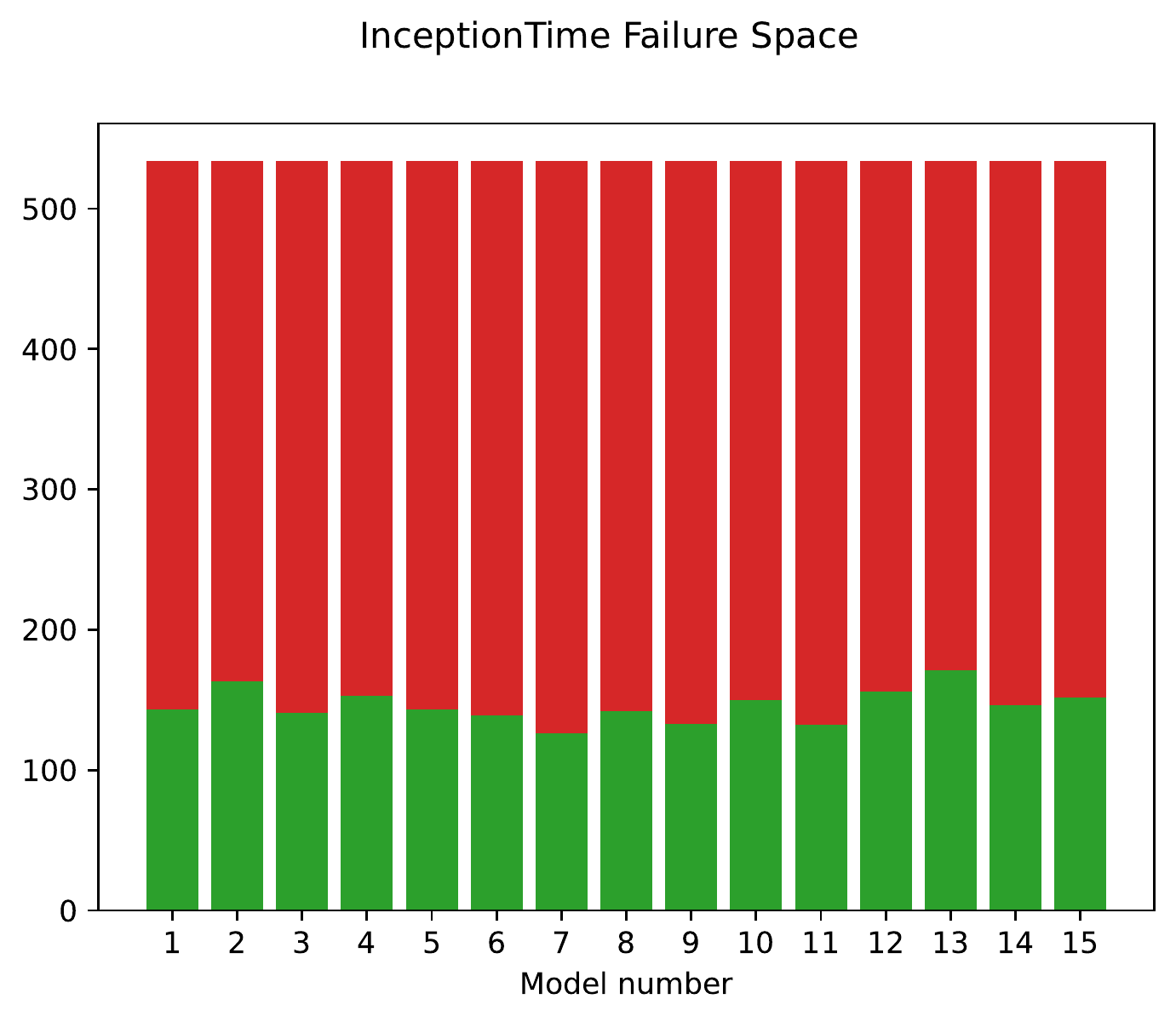}
\caption{The Failure space analysis of InceptionTime in the lower independent fold 0 dataset. The models are numbered as follows: 1:XceptionTime, 2: ResNet, 3: LSTM FCN, 4: BiLSTM FCN, 5:LSTM, 6:BiLSTM, 7:MLSTM FCN, 8:BiMLSTM FCN, 9:FCN, 10:ResCNN, 11:XCM, 12: Optimized InceptionTime, 13: Optimized LSTM, 14: Optimized LSTM FCN, 15:Optimized MLSTM FCN.}
\label{fig:Failure Space InceptionTime}
\end{figure}

Since InceptionTime is the best performing model, next we look into how the failure space of InceptionTime differs from the failure space of the other models. Each bar (green + red combined) in Figure~\ref{fig:Failure Space InceptionTime} represents the total number of incorrect predictions made by the InceptionTime algorithm in the L-WI fold 0 dataset. The green proportion of the bar shows all of the elements where the given model predicted correctly.
As seen, approximately $1/4$'th of all elements which are predicted incorrectly by InceptionTime, are predicted correctly by any of the other given models. In other words, we conclude that the models are failing, or incorrectly predicting, in different places. This leads us to believe that by using an ensemble learning technique, to combine these models, we may be able to reduce the failures outputted. This will be explored in greater detail below. It is also interesting to note that the bar representing the optimized version of InceptionTime, is approximately the same height as the other bars. This may be surprising as one might think that these models would fail in more similar cases and correctly predict in similar cases (with small variances due to optimizations), due to their similar architectures. This indication shows that when exploring ensemble methods below, not only may it be beneficial to use models of differing architectures to help reduce failures spaces, but also include similar architectures as well.

After examining the failure spaces of each of the models, it was found that the models were failing in different places. What this means is that for a given test case, a portion of the 16 models can fail, but oftentimes others will succeed. From here we hypothesized that there may exist some combination of models, which can work together to make predictions more accurately than individual models.

\subsubsection{Binary reclassification}
The first attempt we made at ensemble learning was a binary reclassification model as seen in \cite{7355372} or \cite{porwal_carranza_hale_2006} with mixed results. The first step is to combine the classes of commonly mixed up letters, such as (“o”,”O”), (“s”,”S”) or (“i”,”j”). This can be explored as a problem on its own, as it may be valuable in certain situations to view (“o”, “O”) as the same letter for example (which they are). For the original problem, however, each letter should be viewed as a separate class. After our initial classification, where some letters are combined with each other, there must come a binary classification of these combined letters in order to finish the classification process. The binary reclassification process begins with a combined multiclass classification, before addressing the combined classes by reclassifying them in a binary manner. This acts as a divide and conquer type approach.
Unfortunately the binary reclassification attempt did not result in increased accuracies through our attempts.

\subsubsection{Plurality voting }
Through our failure space analysis, we wanted to try to find a way to combine multiple models to make one single prediction. The first attempt at this is plurality voting, where each model will make a prediction on a piece of data and vote on the predicted class. The final output class is the class which receives the most votes \cite{zhou_2012}. A tie in plurality voting is broken arbitrarily. This was problematic as we wanted a more systematic way to break ties. A lexicographic ordering scheme to break ties was considered. 
This method produced results slightly worse than our best performing models. This however is not unexpected since the best models and worst models all have an equal say in the output. One way to resolve this will be weighted voting and will be explored in the section below. Another resolution is to drop the poorly performing models and only consider the top tier (best 3 models) for the plurality voting. This resulted in an increase of up to $1\%$ on the best performing algorithm for each dataset.

\subsubsection{Weighted voting }
In cases where the individual classifiers are of unequal performance, it can be valuable to give more voting power to the stronger performing models and less voting power to the weaker models \cite{zhou_2012}. As can be seen in Table \ref{tab:results}, we trained 16 models with varying levels of performance and accuracy. In the Plurality voting method, the top 4 best models were considered in the voting process for the best results. For better accuracy, we tried to include more models in a weighted voting method, with lower weights. The models were divided into 3 categories: top, middle, and bottom tier. Out of the 16  models considered, 4 belonged in the top tier (InceptionTime, Optimized InceptionTime, XceptionTime and ResNet), 4 in the middle tier (ResCNN, Optimized LSTM, Optimized MLSTM\_FCN and Optimized LSTM\_FCN) and 8 in the bottom tier (LSTM, BILSTM, MLSTM\_FCN, BIMLSTM\_FCN, FCN, BILSTM\_FCN, LSTM\_FCN,XCM). The weights were established so that the top tier models are considered first and lower tier models are considered only in the events of a tie. The bottom tier of models get a weight of 1 applied to their vote, the middle tier gets a weight of 9 applied to their vote and the top tier gets a weight of 45 applied to their vote. If all 4 bottom tier models agree on a class, their vote is still outweighed by a single middle tier model. Similarly, if all middle tier models agree on a class, their vote is outweighed by a single top tier model. 

\subsubsection{Soft voting}
The models trained above and used in our ensemble models, provide a probability that a given input belongs to each of the letter classes. The class with the highest probability becomes the  predicted class from the model and up until now is all that has been considered. In soft voting, for each class, the probabilities of each prediction class are averaged. The class with the highest average is outputted. Soft voting was implemented, with an improvement on both plurality voting and weighted voting
\subsubsection{Weighted soft voting }
The strongest approach we have found is the weighted soft voting model. This combines the benefits of soft voting, with the benefits of weighted voting. Provided here is a soft model, that gives weights to the probabilities depending on the quality of model. This model, using the same weights as described above was the best performing of the ensemble methods tried and yielded between a $0.91\%$ and $2.98\%$ increase 

\subsubsection{Ensemble learning in practice}

As seen above in Table~\ref{tab:results}, the accuracy of the ensemble learning techniques are competitive with or improved upon existing methods. An increase in accuracy when using ensemble methods is consistent with the literature \cite{CostAndAccuracy,AccuracyMalware,AccuracyHeart,Accuracy}. However, we should explore the impact of implementing a protocol like this in a practical application.

When exploring WI datasets, we found increased accuracies when compared with non-ensemble based techiniques. Training models on WI datasets is performed once. Subsequent inferrence will be made using the model trained. WI trained models tend to generalize better to larger amounts of users when compared with WD models~\cite{UserIndependent}. Since WD models require all users to be part of the training process, with the addition of new users, new models should be trained. It is important to examine the accuracy vs.~training-time trade-off~\cite{TrainingTimetradeoff,CostAndAccuracy} of using ensemble learning models.

Not only do ensemble learning models take longer to train, but they are also larger in size and  require more time to infer predictions. The ensemble learning methods described above, require access to multiple models. This means the size of the ensemble learning method is the combination of the sizes of all of the models which make up the ensemble model. Additionally, in order to make a prediction, a prediction must be made in each of the models, before each of these predictions are used by the ensemble to make one inference. This means that the time required to make an inference is the combination of the times required by each of the individual models, plus some small overhead. Time sensitive predictions may not be the right application for ensemble learning methods for this reason. The time required to make an inference and the size of the models must both be considered when using ensemble models in practice. We hypothesize that user-device based prediction may not be practical with ensemble methods. To use ensemble methods, it may be useful or even required to use server-based predictions as the models can be housed on a server with sufficient storage and computing power.

\section{Further explaining DL models}
\label{s: More Explanations}
It is important to not treat these DL models as black box, but to gain some intuition as to why predictions are being made in these ways. We already provided some explanations regarding failure spaces of the models, which motivated the use of ensemble learning. In this section, we use an interpretation of LIME \cite{ribeiro_singh_guestrin_2016} for time series data, lime-for-time\cite{Lime-For-Time}.  In this interpretation, the time series is divided into $20$ “slices” and the importance of each slice is determined by the LIME algorithm. Additionally, since we are working with MTS data, each of the $13$ signals (channels) will be analyzed separately. The top $30$ signals and slices, in terms of importance, will be examined. A green bar, indicating a score greater than $0$, represents that this section of the data has a positive impact on the model output and a red bar, indicating a score less than 0, represents that this section of the data has a negative impact on the model output. The graphs below show all $13$ channels of the raw data with an importance bar overlayed on top of it. The darkness of colour indicates how much influence that slice of the data has on the classification.

\begin{figure}[t]
\centering
\includegraphics[width=0.7\textwidth]{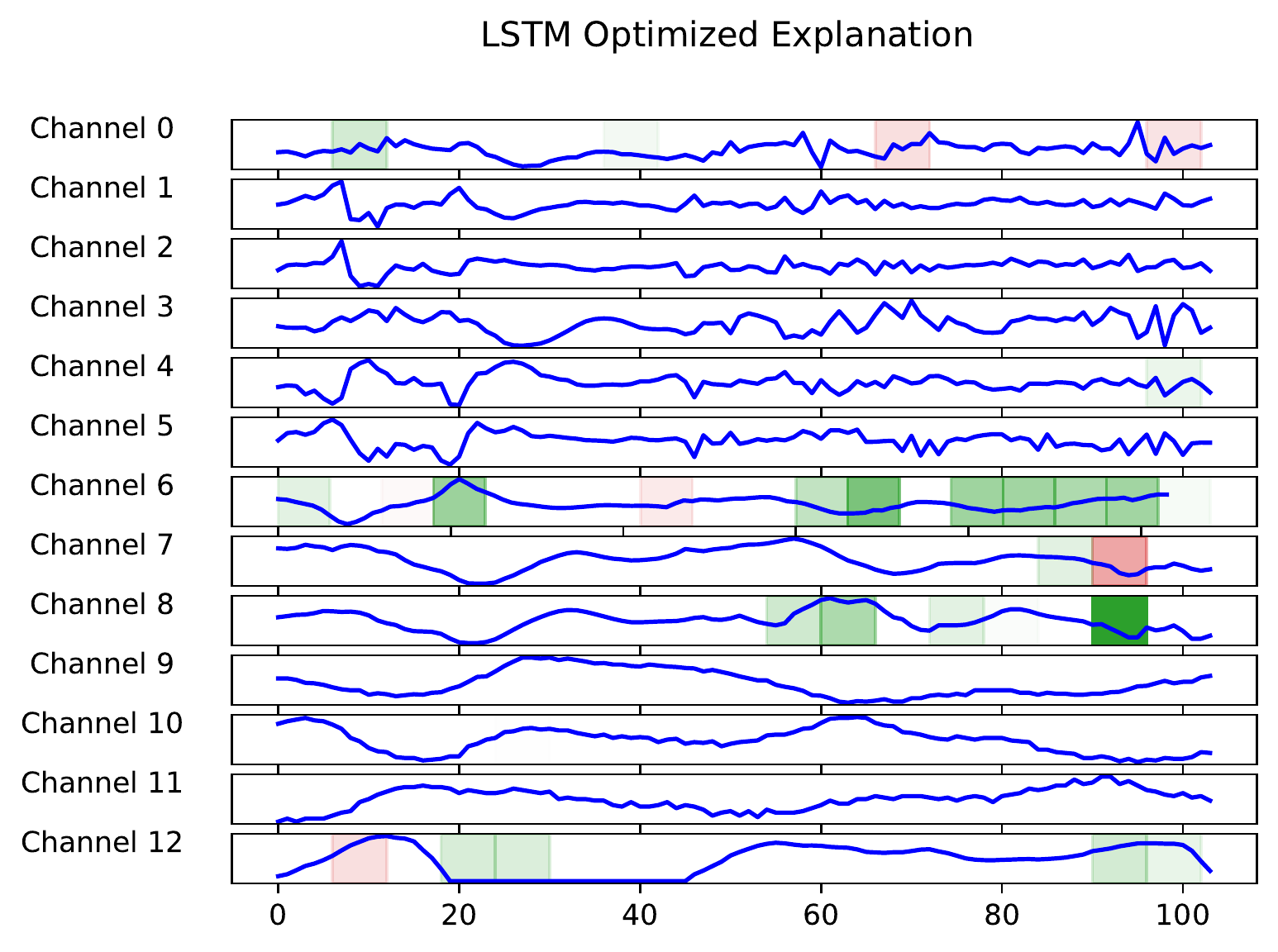}
\caption{Explanation of LSTM Optimized on the both independent fold 0 dataset, for a test letter ``B". Blue lines represent $13$-dimensional data with respect to time.}
\label{fig:LIME LSTM}
\end{figure}

We show two examples of this explanation analysis, to give some backing and explanation as to the predictions being made. The first is an example of where a model predicts correctly and is fairly certain about its prediction. This is shown in Figure~\ref{fig:LIME LSTM} regarding the explanation of the Optimized LSTM model correctly predicting a letter ``B" in the C-WI fold 0 dataset. Optimized LSTM was chosen for this example to provide a different architecture from the second example below, whilst still being a strong performing model. As we can see, the green and red bars are primarily on the 6th, 8th and 12th channel. This means that these channels have more influence (or importance) on the overall prediction. Each channel corresponds to a signal from a sensor and thus different dimension of the MTS.
These results are consistent among most of the models and most of their predictions. 

\begin{figure}[t]
\includegraphics[width=0.5\textwidth]{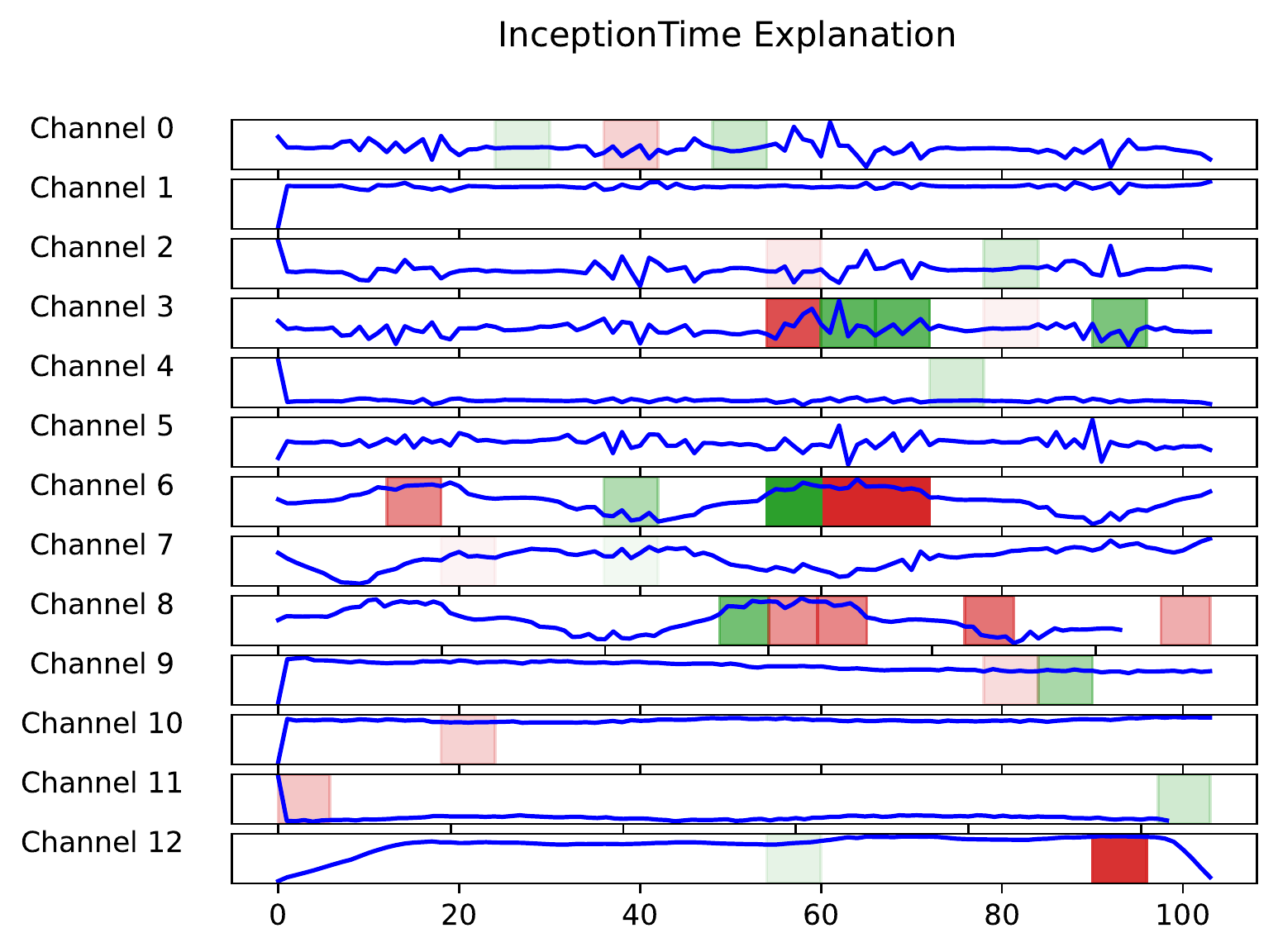}\includegraphics[width=0.5\textwidth]{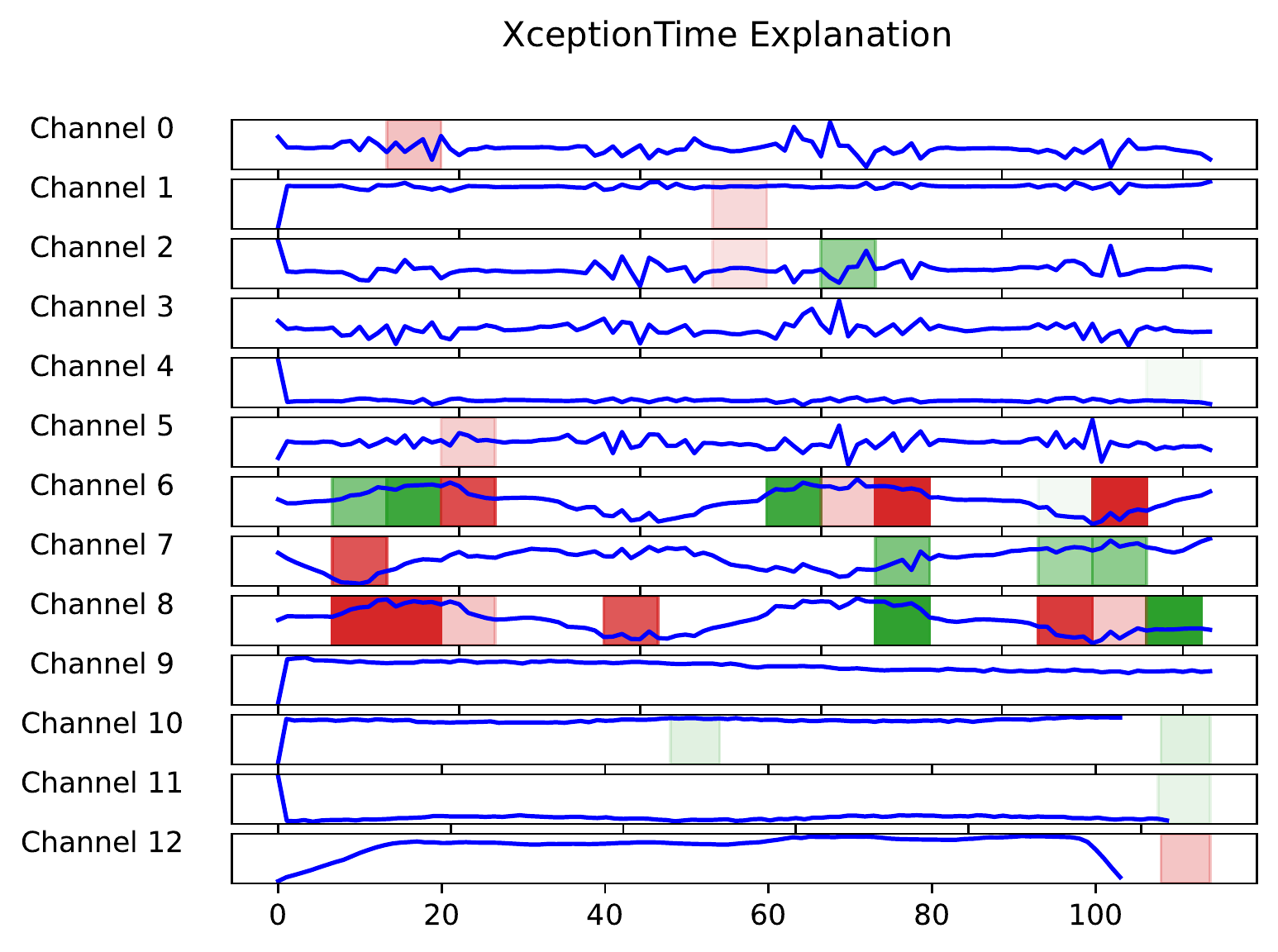}
\caption{The first image on the left is the explanation of a correct InceptionTime prediction of the letter ``M". The second image on the right is the explanation of an incorrect XceptionTime prediction of the same letter. XceptionTime predicts ``A". Blue lines represent $13$-dimensional data with respect to time.}
\label{fig:LIME InceptionTime and XceptionTime}
\end{figure}
The second example we show the explanation of two different (yet both strong) models, which disagree about the output class. This is shown in Figure \ref{fig:LIME InceptionTime and XceptionTime}.  As we can see both of these models inherit influence from different places. One example of this is that XceptionTime puts influence into the first channel for this prediction whereas InceptionTime does not, while InceptionTime puts influence into the 4th channel and XceptionTime does not. Additionally, we see different sections of the 6th, 8th and 12th channels being highlighted. What this tells us is that not only are different models making different predictions from each other, but they are deriving importance from different parts of the MTS. This leads us to believe that by combining these models, we may be able to take advantage of different models inheriting importance from different areas, thus improving the overall ``catchment" of the models chosen.
Along with the failure space analysis, showing that the 16 models fail in different places, these explanations show that the models also value different parts of the signal to have different levels of importance. These two pieces of analysis provide even more motivation on the previous section on ensemble learning, as this provides reason to believe that there may be potential in combining these models to produce a signal output, with better success than any individual model.

\section{Concluding remarks}
\label{s: Conclusion}

We developed various handwriting recognition models on the OnHW-chars dataset \cite{OnHW-2020}. 
Our ML models improved the accuracy of the previously known ML models \cite{OnHW-2020} up to
$23.56\%$ and our DL models (coupled with ensemble methods)
improved the accuracy of the previously known DL models \cite{OnHW-2022} up to $7.01\%$.
We also provided some level of explainability for our models. Our explanations motivated the use of ensemble learning to boost the accuracy of our models and justified its success. Our results can be reproduced and verified via the public repository \cite{impacx_onhw_2022}. Being the most recent and publicly available online handwriting
dataset with some state-of-the-art classifiers trained on it, we have chosen to develop our models on the OnHW-chars dataset. We expect that our techniques would be applicable to other datasets in a more general context.

An interesting future work would be to develop better ensemble methods. For example, one could try to optimize the weights in the voting stage using meta learners or combinatorial approaches \cite{MetaLearning}.
Another approach would be to consider Bayesian model averaging \cite{BayesianModelAveraging}. Finally, it would be interesting to run a deeper analysis on the explainability of models, and to understand the importance of features as this would provide some insight to improve the performance and efficiency of the models.

\section*{Acknowledgments}
We would like to thank Burcu Karabina, Alfred Menezes, Patrick Paul and Pravek Sharma for their valuable feedback and suggestions.
This research has been partially supported by NSERC and the 
National Research Council of Canada's Ideation Fund, New Beginnings Initiative. 

\section*{Author contributions}
The authors of this paper are listed alphabetically and their major contributions are as follows. Hilda Azimi: data preprocessing; optimization of DL models. Steven Chang: data preprocessing; feature extraction; implementation of ML models (Strategy 1). Jonathan Gold: implementation of ML models (Strategy 2) and ensemble learning; explainability of DL models. Koray Karabina: supervision of the project; feature extraction; implementation of ML models (Strategy 1); implementation and optimization of DL models. All of the authors contributed to the writing, reviewing and editing the original and subsequent drafts of this paper.

\bibliographystyle{plain}
\bibliography{references.bib}


\end{document}